\documentclass[lettersize,journal]{IEEEtran}
\usepackage{amsmath,amsfonts}
\usepackage{algorithmic}
\usepackage{algorithm}
\usepackage{array}
\usepackage[caption=false,font=normalsize,labelfont=sf,textfont=sf]{subfig}
\usepackage{textcomp}
\usepackage{stfloats}
\usepackage{url}
\usepackage{verbatim}
\usepackage{graphicx}
\usepackage{cite}
\usepackage{algorithm}
\usepackage{algorithmic}
\usepackage{booktabs}
\usepackage{multirow}
\usepackage{graphicx}
\usepackage{amssymb}
\usepackage{hyperref}

\begin{document}

\title{Ambiguity-Guided Learnable Distribution Calibration for
Semi-Supervised Few-Shot Class-Incremental
Learning}

\author{Fan Lyu,~\IEEEmembership{Member,~IEEE}, Linglan Zhao, Chengyan Liu, Yinying Mei, Zhang Zhang,~\IEEEmembership{Member,~IEEE},\\ Jian Zhang, Fuyuan Hu,~\IEEEmembership{Member,~IEEE}, Liang Wang,~\IEEEmembership{Fellow,~IEEE}
\thanks{
Fan Lyu, Zhang Zhang, and Liang Wang are with the New Laboratory of Pattern Recognition, Institute of Automation, Chinese Academy of
Sciences, 100190, Beijing, China. 

Linglan Zhao is with the School of Electronic Information and Electrical Engineering, Shanghai Jiao Tong University, Shanghai, 200240, China.

Chengyan Liu, Yinying Mei, Jian Zhang and Fuyuan Hu are with the School of Electronics and Information Engineering, Suzhou University of Science and
Technology, Suzhou, 215000, Jiangsu, China and the Jiangsu Industrial Intelligent Low Carbon Technology Engineering Center, 215009, Suzhou, Jiangsu, China.




The corresponding author is Fuyuan Hu (fuyuanhu@mail.usts.edu.cn).
}
}

\maketitle

\begin{abstract}
Few-Shot Class-Incremental Learning (FSCIL) focuses on models learning new concepts from limited data while retaining knowledge of previous classes. 
Recently, many studies have started to leverage unlabeled samples to assist models in learning from few-shot samples, giving rise to the field of Semi-supervised Few-shot Class-Incremental Learning (Semi-FSCIL). 
However, these studies often assume that the source of unlabeled data is only confined to novel classes of the current session, which presents a narrow perspective and cannot align well with practical scenarios.
To better reflect real-world scenarios, we redefine Semi-FSCIL as Generalized Semi-FSCIL (GSemi-FSCIL) by incorporating both base and all the ever-seen novel classes in the unlabeled set. 
This change in the composition of unlabeled samples poses a new challenge for existing methods, as they struggle to distinguish between unlabeled samples from base and novel classes.
To address this issue, we propose an Ambiguity-guided Learnable Distribution Calibration (ALDC) strategy. 
ALDC dynamically uses abundant base samples to correct biased feature distributions for few-shot novel classes.
Experiments on three benchmark datasets show that our method outperforms existing works, setting new state-of-the-art results.
\end{abstract}

\begin{IEEEkeywords}
Few-shot learning, Class-incremental learning, Semi-supervised learning, Distribution calibration.\end{IEEEkeywords}

\section{Introduction}
\label{sec:intro}

\begin{figure}[t]
	\centering

	\includegraphics[width=1.\linewidth]{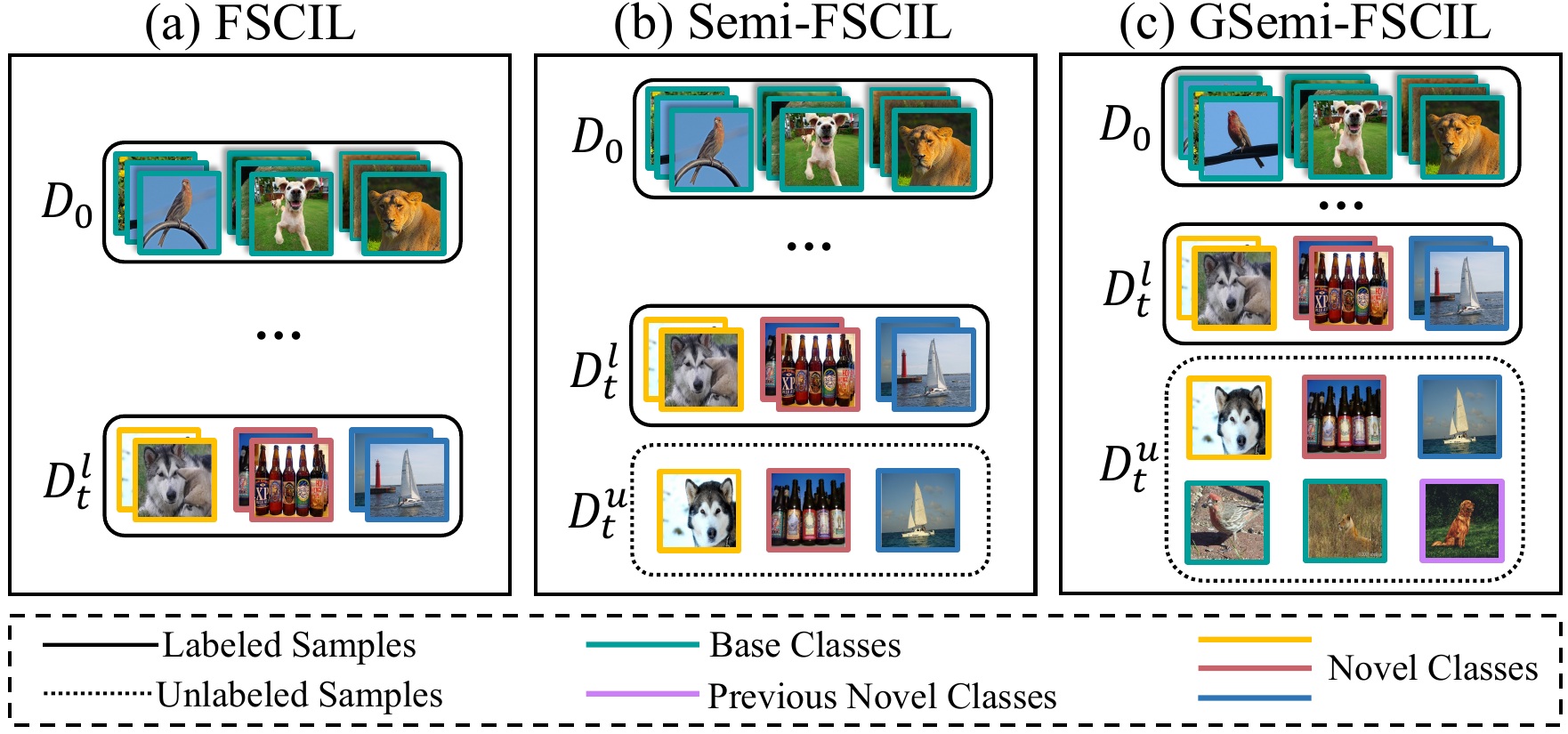}
	\caption{
                    Comparisons between configurations of (a) FSCIL, (b) Semi-FSCIL, and our proposed (c) GSemi-FSCIL. 
                    Best viewed in color and with zoom.
                }
	\label{fig:setting}
\end{figure}

Deep Learning (DL) has demonstrated remarkable success across numerous applications based on the availability of pre-collected datasets. 
However, real-world scenarios often involve data streams, requiring models to learn new information while retaining prior knowledge, known as Class-Incremental Learning (CIL)~\cite{sun2022exploring,lyu2021multi,chen2022harnessing,lyu2023measuring,chen2023multi,dinghang2024metamask,zhaosafe}. 
Moreover, in cases such as endangered species identification, where data is scarce, Few-Shot Class-Incremental Learning (FSCIL)~\cite{tao2020few, zhang2021few, zhou2022forward, Ji2023MCNet, BilateralKD, liu2024cala}, as shown in Fig.~\ref{fig:setting}(a), has been developed. 

FSCIL addresses the challenge of incrementally learning novel classes with limited data while maintaining performance on previously learned classes. The task is organized into multiple sessions ($0 \sim t$), starting with a base session ($0$), where the model is trained on a relatively large dataset with sufficient samples across multiple classes to establish a strong foundation. This is followed by incremental sessions ($1 \sim t$), where only a few labeled samples (typically 1$\sim$5 per class) from novel classes are sequentially introduced, and the data from previously seen classes becomes unavailable, simulating real-world constraints.

Recent studies in Semi-supervised FSCIL (Semi-FSCIL), as shown in Fig.~\ref{fig:setting}(b), use unlabeled samples to enhance FSCIL~\cite{strongbaseline, UaD-CE, Semi-FSCIL, Us-KD}, but typically focus only on novel class (session $t$) samples, which is unrealistic. To better adapt the model to real-world scenarios, we propose a new Generalized Semi-FSCIL setting by incorporating the base and all the ever-seen novel classes (session $1 \sim t$) into the unlabeled dataset, as shown in Fig.~\ref{fig:setting}(c). Due to the changes in the composition of the unlabeled samples, existing methods struggle to distinguish the unlabeled samples from the base and novel classes.
To solve this, we propose an Ambiguity-guided Learnable Distribution Calibration (ALDC) strategy. Our method identifies high-ambiguity samples and dynamically adjusts the selection standard to improve pseudo-labeling accuracy and reduce misclassification.
\begin{figure*}[htbp]
	\centering

	\includegraphics[width=0.95\linewidth]{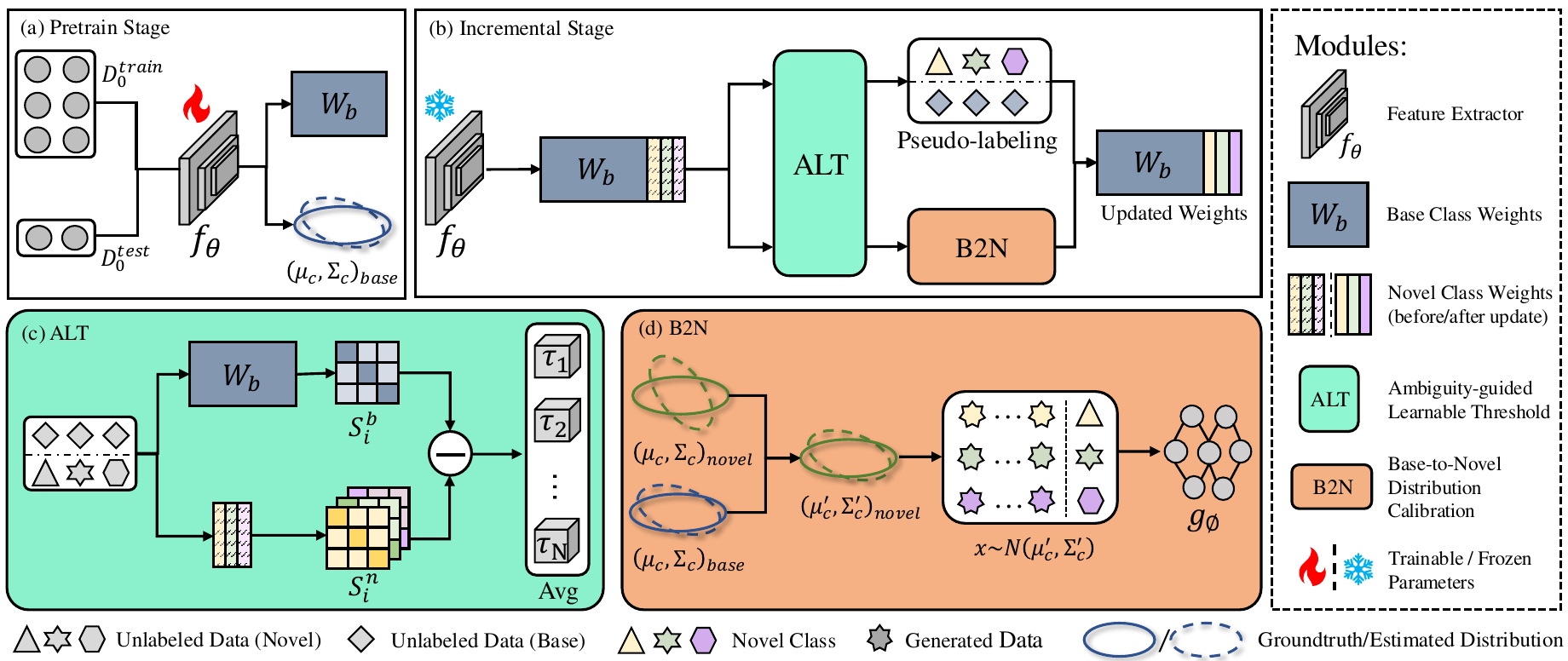 }
	\caption{An overview of our method for GSemi-FSCIL, where (a) and (b) illustrate the base and the incremental stages, respectively. Best viewed in color.}
	\label{fig:framework}
\end{figure*}

Specifically, we design a high-ambiguity unlabeled sample selection strategy to help generate the feature distribution for novel classes. First, we select high-confidence samples from the unlabeled dataset, either from the base or novel classes and assign pseudo-labels for distribution calibration. Next, we screen out high-ambiguity samples, which are difficult to classify as they resemble both base and novel classes. We then calibrate their feature distribution from base to novel. Using this calibrated distribution, we randomly generate enough training samples to build a more comprehensive sample relationship, correcting biased prototype features.

The main contributions of our work are as follows:


\begin{enumerate}
\item We redefine the Semi-supervised Few-Shot Class-Incremental Learning (Semi-FSCIL) benchmark by including both base and novel classes into the unlabeled samples and propose a more challenging and realistic new benchmark: Generalized Semi-FSCIL (GSemi-FSCIL). 

\item We propose an Ambiguity-guided Learnable Threshold (ALT) and a Base-to-Novel (B2N) distribution calibration method to address the confusion between old and novel classes in unlabeled datasets. ALT dynamically selects unlabeled samples and employs a two-branch structure to enhance pseudo-label accuracy, while B2N leverages high-ambiguity samples by calibrating their distribution using similar base class features. 

\item We experimentally show that existing Semi-FSCIL methods degrade in performance under our proposed setting. Comprehensive evaluation on benchmark datasets proves the superiority of our approach over SOTA methods.
\end{enumerate}

\section{Related work}

\textbf{Few-Shot Class-Incremental Learning}.
FSCIL\cite{tao2020few}, \cite{zhang2021few}, \cite{zhou2022forward}, \cite{Ji2023MCNet}, \cite{BilateralKD}, \cite{Fantasy} has emerged to address the challenge of handling few-shot samples within an incremental learning context. 
CEC\cite{zhang2021few} introduces an additional graph model to facilitate the propagation of context information among classifiers, aiding in model adaptation.
FACT\cite{zhou2022forward} efficiently integrates new classes with forward compatibility while simultaneously preventing the forgetting of previously learned ones. 
Bilateral Distillation\cite{BilateralKD} proposes a novel distillation structure that draws knowledge from two different complementary teachers. 
These methods \cite{zhou2022few, Xu2023MFS3, zhu2021self,AsyCon} represent innovative approaches to tackle the challenges of FSCIL, offering different strategies to balance adaptation to new classes and retention of old knowledge. 
However, neither of these methods systematically takes into account the role of unlabeled samples, which are free available in real-world scenarios.

\textbf{Semi-Supervised Learning.}
Semi-Supervised Learning (SSL) has become a pivotal approach for leveraging limited labeled data alongside abundant unlabeled samples. Existing SSL methods can be broadly categorized into consistency regularization \cite{Vat, mixmatch}, pseudo-labeling \cite{transductive} and transductive learning \cite{transductivelearning}. Consistency-based methods enforce output stability under input perturbations, often requiring heavy domain-specific augmentations.  Pseudo-labeling methods generate synthetic labels for unlabeled data using model predictions \cite{pseudo} or graph-based assignments \cite{label}. Transductive models, such as TSVMs \cite{tsvms}, incorporate test data during training to directly optimize performance on specific target instances.

\textbf{Semi-Supervised Few-Shot Class-Incremental Learning.}
The existing Semi-FSCIL methods can be divided into two types: single-stage incremental methods~\cite{strongbaseline} and multiple-stage incremental methods~\cite{UaD-CE, Semi-FSCIL, Us-KD}. While single-stage incremental methods offer a more balanced composition of unlabeled samples, they split the learning process into base and single incremental stages, which may not fully meet FSCIL needs. Multiple-stage incremental methods, however, only use unlabeled samples from the current novel classes, lacking diversity by excluding samples from the base and all ever-seen novel classes, which limits their applicability to real-world scenarios.
\section{Generalized Semi-FSCIL Configuration}

 According to standard FSCIL Settings, our GSemi-FSCIL comprises a base session and incremental sessions. During the base session, a sufficient number of labeled samples are provided for training. Conversely, in incremental sessions, the model must adapt to novel classes using only a limited number of samples. Let \(\mathcal{D} = \{\mathcal{D}_{t}\}_{t=0}^T\), where \(\mathcal{D}_{t} = \{(x_i, y_i)\}_{i=0}^{N \times K}\) represents the training samples at session \(t (t \geq 0)\), where \(x_i\) and \(y_i\) denote the \(i\)-th sample and its corresponding label, respectively. \(y_i\) $\in$ \(C_t\) and \(C_t\) denotes the $t$-th session category set. \(D_0\) represents the large-scale base dataset used in the base session, while subsequent datasets are novel few-shot datasets. In the \(t\)-th session (\(t > 1\)), the novel dataset is denoted as \(D_{t} = D_{t}^{l} \cup D_{t}^{u}\), where \(D_{t}^{l}\) and \(D_{t}^{u}\) represent the labeled and unlabeled training data, respectively. Compared to Semi-FSCIL, where the label space \(C_{t}^{u}\) = \(\{C_t\}\), our proposed GSemi-FSCIL \(C_{t}^{u}\) = \(\{C_0 \cup C_1 \cup \cdot\cdot\cdot \cup C_t\}\). As shown in Fig.~\ref{fig:setting}, the unlabeled training data \(\mathcal{D}^{u}_{t} = \{(x_i)\}_{i=1}^{M}\) comprises unlabeled samples, sourced from either base class samples not used in the base session or novel class samples not used from the first session to the current session. 

\section{Method}

As shown in Fig.\ref{fig:framework}, the proposed Ambiguity-guided Learnable Distribution Calibration (ALDC) method mainly consists of three stages. In the pre-training stage (a), the feature extractor is trained to retain knowledge of the base classes. In the incremental stage, we propose a dual-branch Ambiguity-guided Learnable Threshold (ALT) selection structure (b) to filter the unlabeled samples. On the one hand, for unlabeled samples with high confidence, we directly use them to update the classifier weights. On the other hand, for unlabeled samples that are difficult to distinguish between base and novel classes, we employ a Base-to-Novel (B2N) distribution calibration module to calibrate the feature distribution of the few-shot classes (c).

The pseudo-code for these methods is outlined in Algorithm\ref{alg:S2FSCIL}.

\subsection{Ambiguity-guided Learnable Threshold.} 
In our proposed GSemi-FSCIL configuration, we change the composition of the unlabeled sample dataset in each incremental stage. Specifically, the sources of the unlabeled samples can be divided into three categories: the base session, the current incremental session, and the previous session 1 to $t$-1. Due to this change in the composition of the unlabeled samples, existing Semi-FSCIL methods face difficulties in distinguishing between the base and novel class samples, resulting in performance degradation. To address this issue, we define samples that exhibit strong similarity to both the base and novel classes as high-ambiguity samples.


To effectively filter out these high-ambiguity unlabeled samples, we propose a dual-branch Ambiguity-guided Learnable Threshold (ALT) selection structure to screen all the unlabeled samples.
Specifically, the features of unlabeled samples extracted by the model are compared with the base class weights and the novel class weights respectively, yielding similarity scores \(S^{b}\) and \(S^{n}\). The class subscript with the maximum confidence is selected from these scores, and the ALT between the maximum confidence of base and novel classes is calculated:
\begin{equation}
    S^{b, n}_{i} = \max (\cos (f_{\theta}(x_i),   w^{b, n})), 
    \label{eq:1}
\end{equation}
where $\cos (\cdot)$ denotes the cosine similarity, $ w^{b, n}$ denotes the base and novel class weights.
\begin{equation}
    \tau = \frac{1}{N_u} \sum_{i=1}^{N_u} \left| S^{b}_{i} - S^{n}_{i} \right| + m,
\label{eq:4}
\end{equation}
where $N_u$ denotes the number of unlabeled samples, $m$ denotes the smoothing coefficient. 
\subsection{Base-to-Novel Distribution Calibration.}
For high-ambiguity samples, we believe there is a potential correlation between the associated base class and novel class. Due to the scarcity of samples in few-shot tasks, the class distribution statistics learned by the model may become biased. However, the base classes, supported by a large number of samples, can provide valuable knowledge to support the learning of few-shot classes. 

To this end, we propose a Base-to-Novel (B2N) distribution calibration module, which leverages the rich knowledge learned from the base classes to calibrate the distribution of the few-shot classes, thereby enhancing the model's learning capability in few-shot tasks. 
Using the distribution from the base class, we transfer the mean and covariance, estimated to be more similar to the novel class based on the cosine similarity between the base and novel classes. We calculate the statistical features of each class using the following formula:
\begin{equation}
    \begin{aligned}&\mu_c=\frac{1}{N_{c}^t}\sum_{i=1}^{N_{c}^t}f_{\theta}(x_i),\\
    &\Sigma_c=\frac1{N_{c}^t}\sum_{x_j\in D_{t}}(f_{\theta}(x_j)-\mu_c)(f_{\theta}(x_j)-\mu_c)^T,\end{aligned}
\label{eq:6}
\end{equation}
we perform distribution calibration~\cite{freelunch} for the novel classes using the statistical features of the base classes as follows:
\begin{equation}
{\mu}_{c}^{\prime}=\frac{\sum_{{i}\in{D}_0}{\mu}_i+ {{{\mu}_c}}}{N_{c}^t+1},{\Sigma}_{c}^{\prime}=\frac{\sum_{i\in{D}_0}{\Sigma}_{i}+{\Sigma}_{c}}{N_{c}^{t}+1}+\alpha,
\label{eq:7}
\end{equation}
having a set of calibrated feature distributions in a target task, we generate a set of feature vectors with labels by sampling from the calibrated Gaussian distributions $x\sim N({\mu}_{c}^{\prime},{\Sigma}_{c}^{\prime})$.

Using more accurate statistical information generated from Base-to-Novel, we can generate additional few-shot samples to augment the training set.
Specifically, for each incremental session, we generate $N_{u}$=$\{25, 50, 75, 125\}$ unlabeled samples. For fairness, we ultimately select 50 unlabeled samples for comparative experiments. Taking the mini-ImageNet dataset as an example, these 50 samples are divided into 5 groups, corresponding to the 5 novel classes in the current session. This extends our training set from an $N$-way $K$-shot setting to an $N$-way ($K$+10)-shot setting, where $N$ represents the number of novel classes, and $K$ represents the number of samples per class. For each session, we use the augmented training set to retrain and update the classifier weights.

\begin{algorithm}
\caption{Semi-supervised Few-Shot Class Incremental Learning}\label{alg:S2FSCIL}
\textbf{Input}: $D_1$, $D_2$, ..., $F()$, session $n$, initialized threshold $t$ \\
\textbf{Output}: $F()$ that can classify all seen categories so far

\begin{algorithmic}[1]
\FOR{each $i$ in $n$}
    \IF{$n = 1$}
         \STATE Train backbone, classifier by $D_1$ 
         \STATE Calculate and store $\mu_{\text{base}}$ and $\sigma^2_{\text{base}}$
    \ELSE
        \STATE Initialize novel weight by labelled $D_n$
        \STATE Calculate similarity $S_{\text{base}}$ and $S_{\text{novel}}$ using Eq.~\eqref{eq:1}
        \IF{$| S_{\text{base}} - S_{\text{novel}} | > t$}
            \STATE $\text{Max}_{\text{index}} \rightarrow$ assign pseudo-labels to unlabeled data
        \ELSE
            \STATE Calibrate base-to-novel distribution using Eq.~\eqref{eq:7}
        \ENDIF
        \STATE Update threshold $t$ via Eq.~\eqref{eq:4}
    \ENDIF
    \STATE Update $\mu_{\text{novel}}$ and $\sigma^2_{\text{novel}}$
    \STATE $\mu, \sigma^2 \rightarrow G_{\text{sample}}$
    \STATE Update classifier weights using labeled and unlabeled data
\ENDFOR
\end{algorithmic}
\end{algorithm}

\section{Experiment}

\begin{table*}[ht]
\small
\centering
\caption{Comparison with the state-of-the-art on mini-ImageNet dataset. “Avg” denotes the average accuracy across all sessions.}\label{lab:mini_compare}
\renewcommand{\arraystretch}{1.0} 
\resizebox{1\linewidth}{!}{
 \begin{tabular}{c l cccccccccccc cc  }
  \toprule
   \multirow{2.5}{*}{Task} & \multirow{2.5}{*}{Method} &\multicolumn{9}{c}{Accuracy in each session(\%)}  & \multirow{2.5}{*}{Avg} & \multirow{2.5}{*}{Improve}\\
  \cmidrule(lr@{0.5em}){3-11}
  &\textbf{} & 0 &1  &2&3 &4&5  &6&7&8& & \\ 
  \midrule              

  \multirow{7}{*}{FSCIL} 
  & iCaRL~\cite{rebuffi2017icarl}              &61.31 & 46.32& 42.94& 37.63& 30.49& 24.00& 20.89& 18.80& 17.21& 33.29 &  \textbf{+36.48}  \\
  
  & TOPIC~\cite{tao2020few}              &61.31 &50.09& 45.17& 41.16& 37.48& 35.52& 32.19& 29.46& 24.42 &39.64 &  \textbf{+30.13}  \\
  
  & CEC~\cite{zhang2021few}                &  72.00& 66.83 & 62.97 &59.43 &56.70 &53.73 &51.19 &49.24 &47.63 &57.75  &  \textbf{+12.02} \\
  
  & FACT~\cite{zhou2022forward}                & 72.56& 69.63& 66.38& 62.77& 60.60& 57.33& 54.34& 52.16& 50.49  & 60.77 & \textbf{+7.00} \\
  
  & LIMIT~\cite{zhou2022few}                &  72.32 &68.47 &64.30 &60.78 &57.95 &55.07 &52.70 &50.72 &49.19&  59.05 &\textbf{+10.72}  \\
  
  & MCNet~\cite{Ji2023MCNet}                & 72.33& 67.70& 63.50& 60.34& 57.59& 54.70& 52.13& 50.41& 49.08  & 55.13 &  \textbf{+14.64} \\
  
  \midrule
  
  \multirow{6}{*}{Semi-FSCIL}
 & SS-iCarl~\cite{Us-KD} &62.98 &51.64 &47.43 &43.92 &41.69 &38.74 &36.67 &34.54 &33.92  &43.50 &\textbf{+26.27}\\
 & SS-NCM-CNN~\cite{Us-KD} &62.98 &60.88 &57.63 &52.80 &50.66 &48.28 &45.27 &41.65 &40.51  &51.26 &\textbf{+18.15}\\

   &Semi-CEC~\cite{zhang2021few} & 72.00 & 65.14 & 61.71 & 58.35 & 56.04 & 52.25 & 50.94 & 49.68 & 46.31 & 56.94 &\textbf{+12.83} \\

   &Semi-Limit~\cite{zhou2022few} & 72.32 & 67.56 & 63.14 & 59.53 & 56.97 & 54.23 & 50.39 & 49.06 & 47.89 & 57.89 &\textbf{+11.88} \\
  
  & Us-KD~\cite{Us-KD} & 72.35 & 67.22 & 62.41 & 59.85 & 57.81 & 55.52 & 52.64 & 50.86 & 50.47 & 58.83 &\textbf{+10.94} \\
  
  & UaD-CE~\cite{UaD-CE} &72.35 & 66.91 & 62.13 & 59.89 & 57.41 & 55.52 & 53.26 & 51.46 & 50.52 & 58.84 &  \textbf{+10.93} \\
  
  \midrule
  \multirow{6}{*}{GSemi-FSCIL}
  & SS-iCarl$^{*}$~\cite{Us-KD} &62.98 &50.03 &46.33 &42.19 &40.56 &37.63 &35.76 &33.23 &32.69  &42.38 &\textbf{+27.39}\\
  &Semi-CEC$^{*}$~\cite{zhang2021few} & 72.00 & 64.36 & 60.57 & 56.98 & 55.41 & 51.36 & 49.74 & 47.35 & 45.21 & 55.89 &\textbf{+13.88} \\
   &Semi-Limit$^{*}$~\cite{zhou2022few} & 72.32 & 66.85 & 62.33 & 58.49 & 54.63 & 53.20 & 49.97 & 48.36 & 46.59 & 56.97 &\textbf{+12.80} \\
  & Us-KD$^{*}$~\cite{Us-KD} &72.35 &64.33 &59.38 &56.41 &54.16 &53.37 &50.16 &48.33 &46.54 &56.11 &\textbf{+13.66}\\
  & UaD-CE$^{*}$~\cite{UaD-CE} &72.35 & 65.43 & 60.63 & 58.27 & 56.13 & 53.07 & 51.44 & 49.96 & 47.10 & 58.26 &  \textbf{+11.51} \\
  & ALDC (ours)   &  \textbf{82.45}  &  \textbf{78.85}  &  \textbf{76.75}  &  \textbf{72.87}  &  \textbf{69.98}  &  \textbf{64.54}  &  \textbf{64.66}  &  \textbf{60.19}  &  \textbf{58.17} & \textbf{69.77}\\
  \toprule

\end{tabular}}
\end{table*}

\subsection{Major Comparison}
The comprehensive results presented in Table~\ref{lab:mini_compare} illustrate that our approach consistently surpasses the current SOTA methods, including FACT~\cite{zhou2022forward}, Us-KD~\cite{Us-KD}, and UaD-CE~\cite{UaD-CE}, on the mini-ImageNet dataset. Across all sessions, our method demonstrates superior performance compared to other approaches, outperforming the SOTA method by 10.93\% in Semi-FSCIL and by 11.51\% in GSemi-FSCIL.
Fig.~\ref{fig:compare} (a-b) illustrates the accuracy of each session on the CIFAR-100 and CUB200 datasets compared to FSCIL and Semi-FSCIL. 
\subsection{Ablation Study}
\textbf{The ablation of each component in ALT.} We analyze the impact of each ALDC component on mini-ImageNet, as shown in Table~\ref{lab:Abl_mini}. The results indicate that assigning pseudo-labels solely based on confidence biases weight updates, leading to degraded performance. Discarding high-ambiguity samples results in the loss of key information and increases overfitting, while a static ALT fails to adapt to varying uncertainties. In contrast, dynamic evolved ALT addresses these issues effectively, resulting in improved performance.

\begin{table}[ht]
\Huge
\centering
\caption{Ablation studies on mini-ImageNet dataset. “Baseline” denotes the utilization of the backbone network to directly learn FSCIL tasks. “Drop” signifies that samples with high-ambiguity are discarded. “Static” indicates that the ALT remains fixed. “Dynamic” denotes the ALT is dynamic evolved.}\label{lab:Abl_mini}
\renewcommand{\arraystretch}{1.2} 

\resizebox{1\linewidth}{!}{
 \begin{tabular}{c lccccccccccc cc  }
  \toprule
   & \multirow{2}{*}{Baseline} &\multicolumn{9}{c}{Accuracy in each session(\%)}  & \multirow{2}{*}{Avg} \\
  \cline{3-11}
  &\textbf{} & 0 &1  &2&3 &4&5  &6&7&8& & \\  
  \hline              

  &{Baseline}           & 82.45 & 76.95 
& 72.37 & 68.37 & 65.00 & 61.82 & 58.71 & 56.31 & 54.45 
  &  66.16    \\

 &{Drop} &  82.45 & 78.05 & 75.25 & 71.61 & 68.33 & 64.59 & 60.97 & 58.47 & 56.90 
 & 68.51 \\
  &{Static} &82.45	&78.85	&76.43	&72.39	&69.76	&64.29	&62.17	&59.86	&57.31 &69.27
 \\
 
 \hline
  & {Dynamic} &\textbf{82.45} & \textbf{78.85} & \textbf{76.75} & \textbf{72.87} & \textbf{69.98} & \textbf{64.54} & \textbf{64.66} & \textbf{60.19} & \textbf{58.17} 
   & \textbf{69.77}   \\
  \toprule
  
\end{tabular}}				
\end{table}

\textbf{The ablation of each component in ALDC.} 
We conducted ablation studies on each module of ALDC and presented the prediction results for both base and novel categories in each session. As shown in Table~\ref{lab:Abl_mini_1}, ALT filters a large number of ambiguous samples, preventing the performance of base classes from being affected by these confusing samples and thus achieving stable performance improvements. Meanwhile, B2N alleviates the overfitting problem to some extent, helping the model achieve performance gains in few-shot tasks. By combining ALT and B2N, our ALDC model achieves superior overall performance.

\begin{table*}[htbp]
\small
\centering
\caption{Ablation studies on each component of ALDC on mini-ImageNet dataset. “Baseline” denotes the utilization of the backbone network to directly learn FSCIL tasks. “ALT” denotes the Ambiguity-guided Learnable Threshold. “B2N” indicates the Base-to-Novel Distribution Calibration.}\label{lab:Abl_mini_1}

\resizebox{1\linewidth}{!}{
 \begin{tabular}{c ccclcccccccc cc  }
  \toprule
   &\multirow{2}{*}{Baseline} &\multirow{2}{*}{ALT} &\multirow{2}{*}{B2N} &\multirow{2}{*}{Classes}  &\multicolumn{9}{c}{Accuracy in each session(\%)}  & \multirow{2}{*}{Avg} \\
  \cline{6-14}
  &\textbf{} &\textbf{}&\textbf{}&\textbf{} & 0 &1  &2&3 &4&5  &6&7&8\\
  \hline              

   &\checkmark    & &     &Base & 82.45 & 80.38 
& 75.69 & 72.43 & 69.04 & 65.44 & 61.37 & 59.85 & 56.73 
  & 69.26   \\
 &\checkmark    & &   &Novel  & - & 48.70 
& 46.54 & 44.26 & 42.19 & 41.66 & 40.07 & 39.98 & 38.24 
  &  42.71    \\
 &\checkmark    & &   &ALL   & 82.45 & 76.95 
& 72.37 & 68.37 & 65.00 & 61.82 & 58.71 & 56.31 & 54.45 
  &  66.27    \\

\hline

 &\checkmark    &\checkmark &&Base &  82.45 & 80.69 & 77.14 & 74.37 & 72.12 & 69.80 & 68.44 & 67.31 & 63.96 
 & 72.92 \\
 &\checkmark    &\checkmark & &Novel & - & 50.33 & 50.16 & 47.28 & 46.34 & 43.22 & 41.98 & 40.07 & 39.66 
 & 44.88 \\
 &\checkmark    &\checkmark & &All &  82.45 & 77.13 & 75.35 & 70.17 & 67.54 & 63.26 & 60.18 & 58.33 & 56.00 
 &  67.82 \\

\hline

  &\checkmark    & &\checkmark &Base  &82.45	&79.64	&76.98	&73.24	&68.57	&67.36	&66.23	&65.48	&62.70 &71.41
 \\
  &\checkmark    & &\checkmark &Novel  &-	&52.35	&51.60	&48.75	&47.14	&45.23	&42.90	&40.88	&40.17 &46.13
 \\
  &\checkmark    & &\checkmark &All  &82.45	&77.56	&75.83	&69.34	&66.79	&64.32	&59.78	&58.30	&56.45 &67.87
 \\
 
 \hline
 
  &\checkmark    &\checkmark &\checkmark &Base  &82.45 & 79.69 &77.83 & 76.52 & 74.38 & 71.60 & 69.88 & 68.40 & 65.32 
   &74.01   \\
  &\checkmark    &\checkmark &\checkmark &Novel &- & 55.37 & 53.36 & 50.56 & 47.38 & 46.74 & 43.33 & 41.19 & 40.63 
  & 47.32  \\
     &\checkmark    &\checkmark &\checkmark &All  &\textbf{82.45} & \textbf{78.85} & \textbf{76.75} & \textbf{72.87} & \textbf{69.98} & \textbf{64.54} & \textbf{64.66} & \textbf{60.19} & \textbf{58.17} 
   & \textbf{69.77}   \\

  \toprule
  
\end{tabular}}
\end{table*}

\begin{figure*}[htbp] 	\centering

	\includegraphics[width=1.0\linewidth]{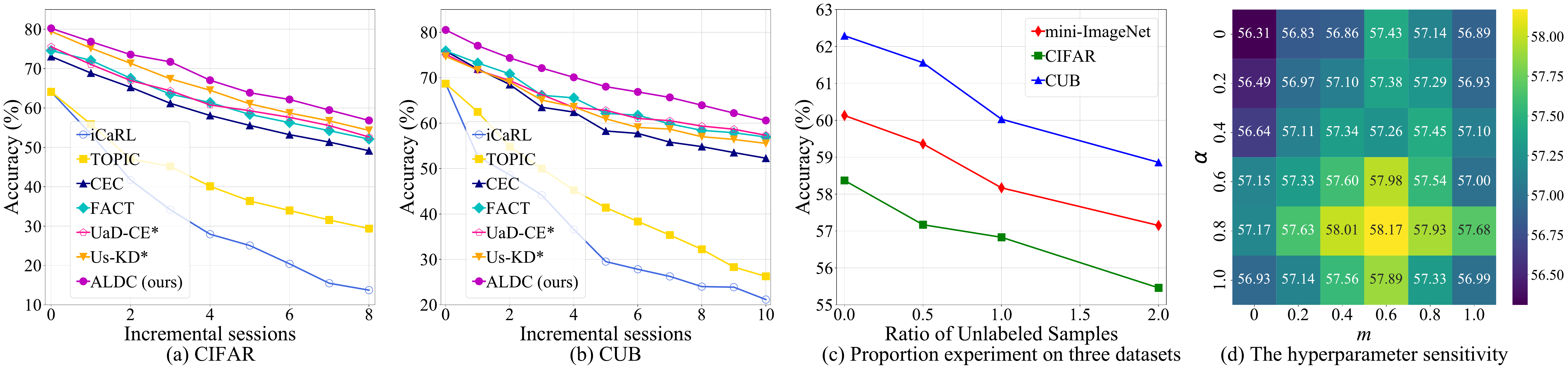}
	\caption{Performance curves of our method compared to  -of-the-art FSCIL and Semi-FSCIL methods on (a) CIFAR100 and (b) CUB, (c) Performance of different methods by varying the sample ratio (base / novel) in the unlabeled set and (d) The hyperparameter sensitivity on min-ImageNet.}
	\label{fig:compare}
\end{figure*}

\textbf{The impact of unlabeled samples.} As shown in Table~\ref{lab:abl_unl_num}, the model achieves the best performance on the mini-ImageNet dataset when 50 unlabeled samples are added in each incremental learning session. To assess how the proportion of unlabeled samples affects performance, we vary this proportion, as shown in Fig.~\ref{fig:compare} (c). Tasks with a higher proportion of novel class samples are less challenging, resulting in lower accuracy when there are fewer novel class samples.

\begin{table}[ht]
\Huge
\centering
\caption{Ablation studies of mini-ImageNet on the number of unlabeled samples. “$N_u$” denotes the number of unlabeled samples.}\label{lab:abl_unl_num}
\renewcommand{\arraystretch}{1.0}  

\resizebox{1\linewidth}{!}{
 \begin{tabular}{l cccccccccccc cc  }
  \toprule
  \multirow{2}{*}{\textbf{$N_u$}} &\multicolumn{9}{c}{Accuracy in each session(\%)}   & \multirow{2}{*}{Avg}\\
  \cline{2-10}
    & 0 &1  &2&3 &4&5  &6&7&8& & \\ 
  \hline              
    25   &  82.45
 &  77.64  &75.38
  &  71.97  &  68.34  &  63.25 &  62.19  & 59.46  &  57.23 & 68.66\\

  50   &  82.45
  &  78.85  &  76.75
  &  72.87  & 69.98  &  64.54  & 64.66  & 60.14  &  58.17 & 69.82\\

75   &  82.45
  &  79.03  & 76.89
  &  72.59  &  70.18  &  65.15  &  64.20  &  60.97  &  58.87 &70.00\\

125  &  \textbf{82.45}
  & \textbf{79.96}  &  \textbf{77.92}
  &  \textbf{72.65} & \textbf{71.56}  &  \textbf{66.74} &\textbf{65.38}   & \textbf{61.19}  &  \textbf{59.33} & \textbf{70.80}\\
  \toprule
\end{tabular}}							
\end{table}

\textbf{The hyperparameter sensitivity anlysis.} We conduct the sensitivity analysis on the two hyperparameters of ALDC and report the accuracy of the last session, as shown in Fig.~\ref{fig:compare} (d). We select $m$ and $\alpha$ from $\{$0, 0.2, 0.4, 0.6, 0.8, 1.0$\}$. It also demonstrates that a proper $\alpha$ can improve the performance of novel classes, which confirms the base classes can benefit the performance of novel classes. Compared with smaller $m$, relatively larger $m$ can smooth the sharpness of ALT.
\subsection{Examples of high-ambiguity samples.}
\begin{figure}[htbp]
    \centering
    \includegraphics[width=1.0\linewidth]{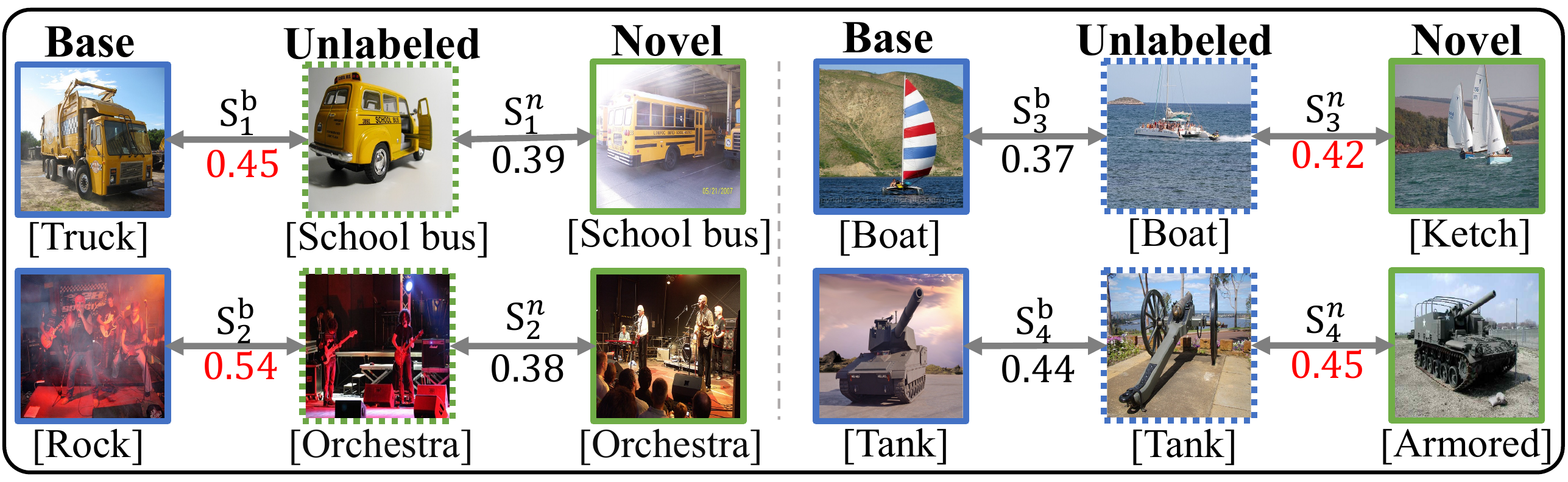}
    \caption{Examples of high-ambiguity samples.}
    \label{fig:Example}
\end{figure}

We have provided examples of high-ambiguity samples from mini-ImageNet dataset. 
As shown in Fig.~\ref{fig:Example}, unlabeled samples show high similarity scores to both base and novel class samples with semantic similarities, making them challenging to distinguish (confusions are highlighted in red).

\vspace{-0.5cm}

\subsection{Visualization}
\begin{figure}[htbp]
    \centering
    \includegraphics[width=1.0\linewidth]{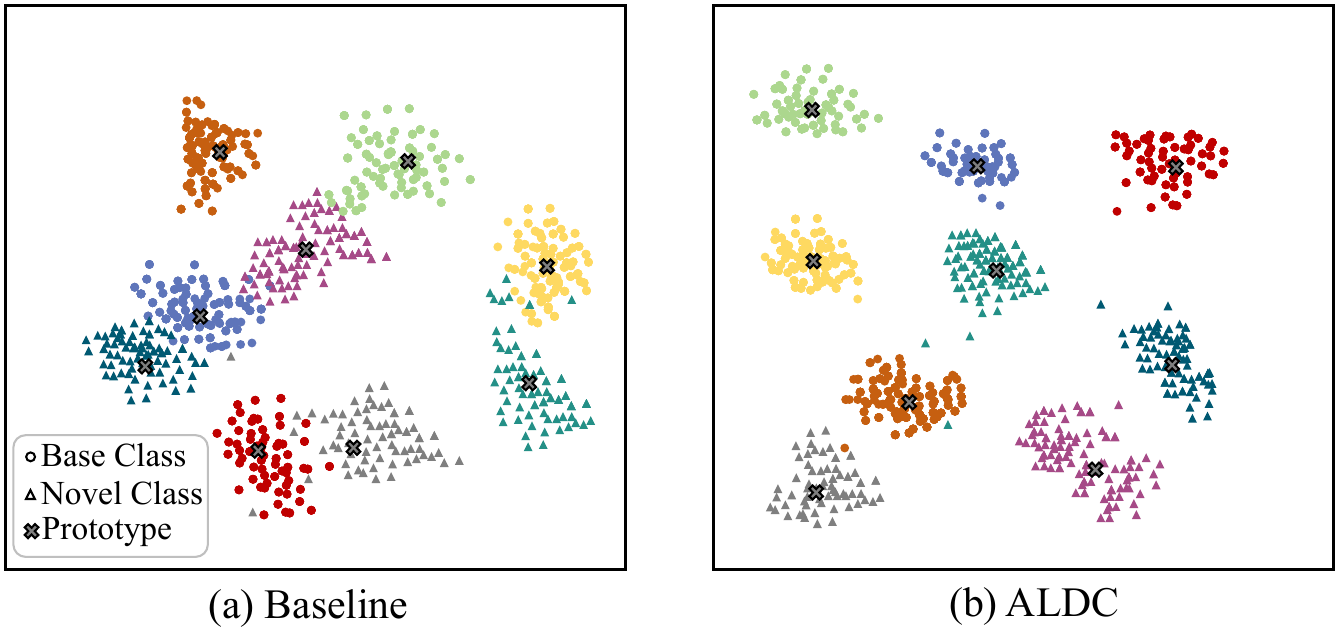}
    \caption{The t-SNE visualization of the mini-ImageNet dataset is illustrated. 
 }
    \label{fig:visual}
\end{figure}

As shown in Fig~\ref{fig:visual}, our t-SNE~\cite{t-sne} visualization of the mini-ImageNet dataset demonstrates that using our ALDC (b) for assigning pseudo-labels produces much clearer decision boundaries compared to relying solely on confidence predictions (a). This improvement highlights the effectiveness of our approach in enhancing class separation, resulting in better clustering and more distinct boundaries in the feature space.

\vspace{-0.3cm}

\section{Conclusion}
In this paper, we propose a novel GSemi-FSCIL configuration that is more adaptable to practical scenarios. We improve the unlabeled sample composition by including both base and novel class samples. To address confusion between these samples, we developed an ALDC method, which incorporates an ALT and B2N to guide the selection of unlabeled samples. Experimental results demonstrate that our model outperforms SOTA methods in both performance and adaptability.

\section*{Acknowledgment}
This work was jointly supported by the National Science and Technology Major Project (2022ZD0117901), the National Natural Science Foundation of China (Nos. 62406323, 62373355, 62476189), the Postdoctoral Fellowship Program of CPSF (No. GZC20232993), and the China Postdoctoral Science Foundation (No. 2024M753496).

\bibliographystyle{IEEEtran}  
\bibliography{main}

\end{document}